\documentclass[letterpaper, 10 pt, conference]{ieeeconf}  %

\IEEEoverridecommandlockouts                              %

\usepackage{graphicx}
\usepackage{amsmath} %
\usepackage{amssymb}  %
\usepackage{bm}
\usepackage{tabularx}
\usepackage{booktabs}
\usepackage[labelformat=simple]{subcaption}

\usepackage{flushend}
\usepackage{stfloats} %
\usepackage{booktabs}
\usepackage{enumerate}   

\newcommand{\beps}{\bm{\epsilon}}
\newcommand{\bp}{\bm{p}}

\usepackage{hyperref}
\hypersetup{
    colorlinks=true,
    linkcolor=blue,
    citecolor=blue,
    filecolor=magenta,      
    urlcolor=cyan}
\usepackage{url}

\newcommand{\rulesep}{\unskip\ \vrule\ }

\usepackage{xspace}
\newcommand{\alg}{\textsc{StackGen}\xspace}%

\makeatletter
\let\NAT@parse\undefined
\makeatother
\usepackage[square, numbers, sort]{natbib}

\title{\LARGE \bf
\alg: Generating Stable Structures from Silhouettes via Diffusion}

\author{
Luzhe Sun$^\ast$ \quad %
Takuma Yoneda$^\ast$ \quad%
Samuel W. Wheeler \quad %
Tianchong Jiang \quad%
Matthew R.\ Walter%
\thanks{L.~Sun, T.~Yoneda, T.~Jiang, and M.R.~Walter are with the Toyota Technological Institute at Chicago (TTIC), Chicago, IL USA, {\tt\small \{luzhesun,takuma,tianchongj,mwalter\}@ttic.edu}. S.W.~Wheeler is with Argonne National Laboratory, Lemont, IL, USA, {\tt\small swwheeler@anl.gov}.}
}

\begin{document}
\maketitle
\thispagestyle{empty}
\pagestyle{empty}

\begin{abstract}
Humans naturally obtain intuition about the interactions between and the stability of rigid objects by observing and interacting with the world. It is this intuition that governs the way in which we regularly configure objects in our environment, allowing us to build complex structures from simple, everyday objects. %
Robotic agents, on the other hand, traditionally require an explicit model of the world that includes the detailed geometry of each object and an analytical model of the environment dynamics, which are difficult to scale and preclude generalization. Instead, robots would benefit from an awareness of intuitive physics that enables them to similarly reason over the stable interaction of objects in their environment.
Towards that goal, we propose \alg---a diffusion model
that generates diverse stable configurations of building blocks matching a target silhouette.
To demonstrate the capability of the method, we evaluate it in a simulated environment and deploy it in the real setting using a robotic arm  to assemble structures generated by the model. Our code is available at \url{https://ripl.github.io/StackGen}.

\end{abstract}

\section{Introduction}
Understanding the physics of a scene is a prerequisite for performing many physical tasks, such as stacking, (dis)assembling, and moving objects. 
Humans can 
intuitively assess and predict the stability of structures through a combination of visual cues, force feedback, and experiential knowledge. 
On the other hand,
robots lack natural multimodal sensory integration and an understanding of intuitive physics.
Robots have traditionally relied upon a world model that includes a representation of the detailed geometry of the objects in the environment and an analytical model of the dynamics that govern their interactions. This dependency poses significant challenges to deploying robotic agents in unprepared environments. 

The ability to compose a diverse array of blocks into a stable structure has a long history as a testbed to study an agent's understanding of object composition and interaction~\citep{battaglia2013simulation,Li2016ToFO,pmlr-v48-lerer16,hamrick2016inferring}. 
While seemingly primitive, this ability
comes with many practical implications such as robot-assisted construction~\cite{helm2012mobile,petersen2019review,ardiny2015construction,gawel2019fully,johns2023framework}, and would serve as a backbone for downstream applications where an agent deals with complex sets of real world objects.

Contemporary approaches to building 3D structures based upon an intuitive understanding of physics utilize the predicted forward dynamics of a scene as part of a planner that combines building blocks into a target structure. This typically involves first training a forward dynamics model that serves as the intuitive physics engine, and then using this model to simulate the behavior of candidate object placements via a form of rejection sampling. Such an approach comes at a high cost as it requires searching through a large space of coordinates and modeling the dynamics for each possible block placement.
\begin{figure}[!t]
    \centering
    \includegraphics[width=\linewidth]{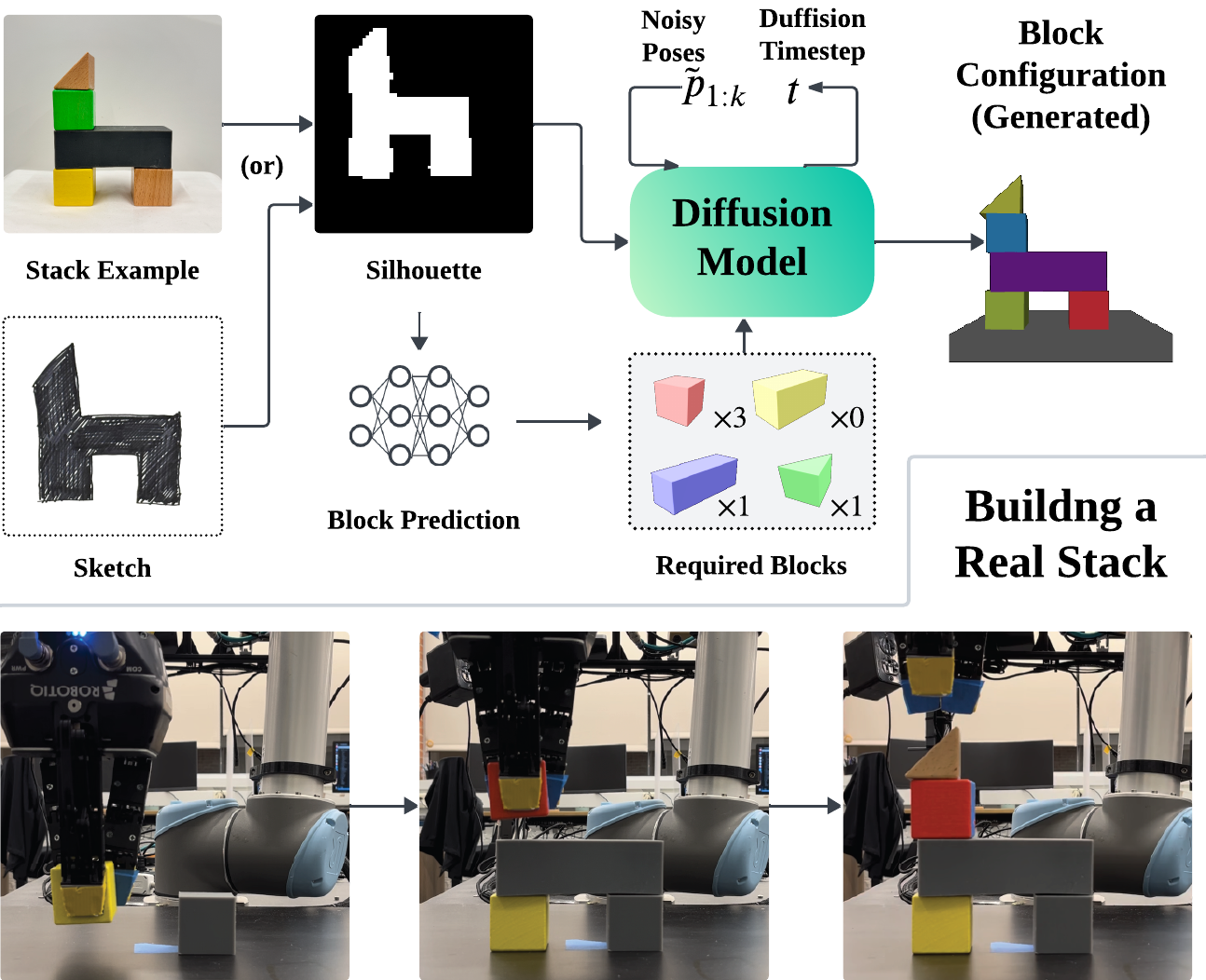}
    \caption{\alg consists of a diffusion model that takes as inputs a target structure silhouette and a list of available block shapes. The model then  generates a set of block poses $\{\hat{\bp}_{1}, \ldots \hat{\bp}_k\}$ that construct a stable structure consistent with the target silhouette. The resulting structure can then be constructed using a robot arm.}
    \label{fig:motivation}
\end{figure}

Rather than training a forward dynamics model,
we 
consider learning and generating a joint distribution over the SE(3) poses of objects composed to achieve a stable 3D structure (Fig.~\ref{fig:motivation}). We condition this distribution on a user-provided specification of the structure, allowing them to control the generation at test time.

Inspired by its success in computer vision~\citep{NEURIPS2021_dhariwal, rombach2021highresolution, pmlr-v162-nichol22a, pmlr-v139-ramesh21a, Ho2022ClassifierFreeDG} and, more recently, robotics domains~\citep{DBLP:conf/iclr/JannerLFTFW19,yoneda23,chi2023diffusion}, 
we employ
conditional diffusion models~\citep{2020Ho-DDPM}, a family of generative models shown to perform well in various
generation domains, in our case for producing
stable 6-DoF object poses. 
Similar in spirit with those approaches that control image generation via spatial information such as sketch or contour~\citep{zhang2023adding}, we ask a user to provide a silhouette that vaguely describes the desired structure, and use it as a conditioning signal. Different from \citet{zhang2023adding}, we simply train a conditional diffusion model built on the Transformer architecture.
We should note that, unlike standard image generation, our approach generates a set of poses. And we aim to generate those that result in a \emph{physically stable} structure.

Our model (\alg) reasons over the 6-DoF pose of different building blocks to realize their composition as part of a stable 3D structure consistent with different user-provided target specifications.
In the following sections,
we describe a Transformer-based architecture that underlies our diffusion model and the procedure by which we generate stable block configurations for training and evaluation. We evaluate the capabilities of \alg through baseline comparisons and as well as real-world experiments that demonstrate its benefits to real-world scene generation with a UR5 arm.

\section{Related Work}
\subsection{Learning Stability from Intuitive Physics}
Similar to our work, several efforts~\citep{Li2016ToFO, pmlr-v48-lerer16, DBLP:conf/iclr/JannerLFTFW19} consider the interaction of relatively simple objects to investigate the notion of intuitive physics.
When considering visual signals for assessing stability, ShapeStacks~\citep{groth2018shapestacks} successfully learned the physics of convex objects in a single-stranded stacking scenario. This is achieved by vertically stacking objects, calculating their center of mass (CoM), and scaling up the dataset to train a visual model via supervised learning, enabling stability prediction prior to stacking. However, calculating the CoM for combinations in multi-stranded stacks proves to be much less straightforward.
Another form of intuitive physics involves the ability to predict how the state of a set of objects will evolve in time, which includes concept of continuity and object permanence. This has motivated the development of benchmarks that measuring models' ability on such tasks called violation-of-expectation (VoE)~\citep{RePEc:nat:nathum:v:6:y:2022:i:9:d:10.1038_s41562-022-01394-8, NEURIPS2019_e88f243b, DBLP:journals/corr/abs-1803-07616, Piloto2018ProbingPK}.
In the context of robotics, \citet{DBLP:journals/corr/AgrawalNAML16} collect video sequences of objects being poked with a robot arm. Using this dataset, they train  forward and inverse dynamics models from pixel input and demonstrate that the model enables the robot to reason over an appropriate sequence of pokes to achieve a goal image. Other work has similarly followed suit~\citep{unsupervised-learning-for-phys-interaction,Finn2016DeepVF}.

\subsection{Diffusion Models for Pose Generation}
Given their impressive ability to learn multimodal distributions, a number of works employ diffusion models~\citep{ho2020ddpm} to learn distributions over the SE(3) poses in support of robot planning~\citep{urain2022se3dif, simeonov2023rpdiff, Yoneda20236DoFSF}. \citet{urain2022se3dif} use conditional diffusion models to predict plausible end-effector positions conditioned on target object shapes for robot manipulation. \citet{simeonov2023rpdiff} use a diffusion model to predict the optimal placements of objects in a scene by modeling the spatial relationships between objects and their environment, identifying target poses for tasks like shelving, stacking, or hanging. Their method incorporates 3D point cloud reconstruction as contextual information to ensure that the predicted poses are both functional and feasible in real-world scenarios.
\citet{structdiffusion2023} and \citet{Xu2024SetIU}  combine large language models with a compositional diffusion model to analyze user instructions and generate a graph-based representation of desired object placements. They then predict object arrangement patterns by optimizing a joint objective, effectively merging language understanding with spatial reasoning. %

\subsection{Automated Sequential Assembly}
Relevant to our data generation procedure, \citet{Tian2023ASAPAS} propose an assembly method (ASAP) that relies on a reverse process of disassembly, where each component is placed in a unique position to guarantee physical feasibility. However, this approach does not account for the potential structural instability that might arise from multiple combinations, since the assembly scenario assumes a one-to-one mapping of components to specific locations.

In contrast, our work addresses the challenge of finding a structure that maintains gravitational stability using only a 2D silhouette through a one-to-many mapping approach. This method ensures that the structural stability is retained and accurately reproduced when transitioning to a 3D environment. Our focus is on the generation and verification of structurally stable block configurations rather than optimizing the assembly sequence. 
Similar to the method of ASAP, which generates step-by-step assembly sequences where intermediate configurations remain stable under gravitational forces, we propose a ``construction by deconstruction'' method that enables scalable data generation by predicting diverse stable configurations, without relying on predefined assembly paths.

In this section, we describe our diffusion-based framework for generating SE(3) poses for blocks that together form a stable structure consistent with a user-provided specification of the scene. We then discuss the procedure for training the model, including an approach to producing a training set that contains a diverse set of stable block configurations.

\section{Method} \label{sec:method}
\subsection{Diffusion Models for SE(3) Block Pose Generation} %

\begin{figure}[!t]
    \centering
    \includegraphics[width=1.0\linewidth]{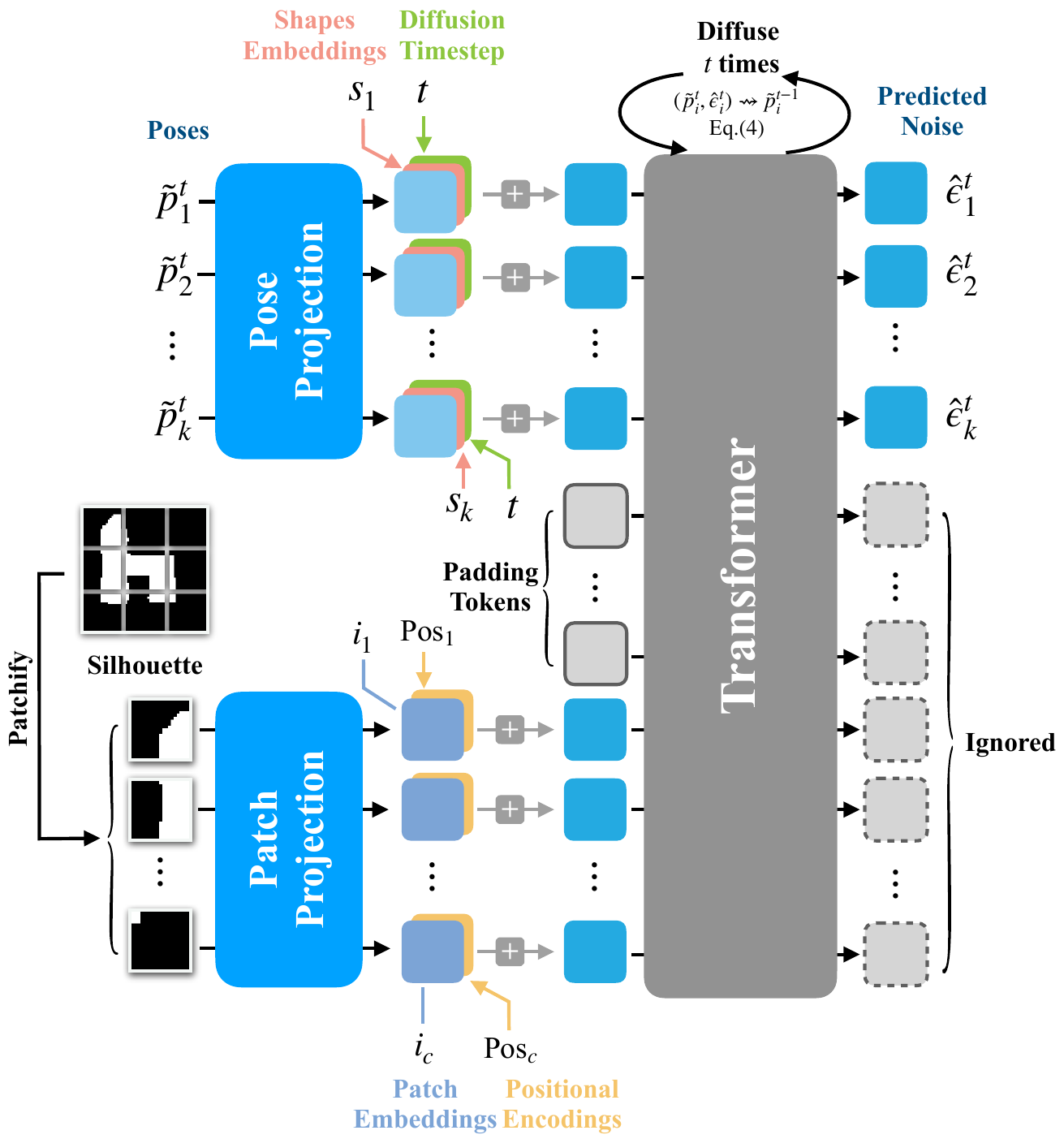} %
    \caption{A visualization of \alg's Transformer-based architecture. We first embed noisy poses $\Tilde{\bm{p}}_{1:k}^t$, shapes $\bm{s}_{1:k}$ and timestep $t$ to same dimension $d$, then add them together to construct object tokens $\in \mathbb{R}^{k\times d}$. Then we patchify the silhouette and embed them to $\bm{i}_{1:c}$, sum up with the positional encoding $\textrm{Pos}_{1:c}$ to construct silhouette tokens $\in \mathbb{R}^{c\times d}$. Then we feed object, padding and silhouette tokens together to our \emph{denoising} Transformer $D_\theta$ (Eq. \ref{eq:denoiser}) in the diagram to predict noise $\hat{\beps}_{1:k}^t$. By running Eq.\ref{eq:one-step-denoise} to get next reverse state $\Tilde{\bm{p}}_{1:k}^{t-1}$.} 
    \label{fig:arch}
\end{figure}

Our model (Fig.~\ref{fig:arch}) generates the SE(3) block poses necessary to create a 3D structure that both matches a given condition (e.g., a silhouette) and is stable. Underlying our framework is a Transformer-based diffusion model that represents the distribution over stable 6DoF poses, without explicitly specifying the number, type, or position of its constituent blocks. In this way, the model employs a reverse diffusion process to produce block poses that collectively form a stable structure. Separately, we train a convolutional neural network (CNN) to predict the number and type of blocks necessary for the construction based on the target silhouette. At test-time, we employ the CNN to predict the block list, and then provide this list and the target silhouette as input to the diffusion model. The diffusion model then samples potential block poses composing a stable structure.

We adopt the denoising diffusion probabilistic model (DDPM)~\citep{ho2020ddpm} as the core framework underlying \alg. A probabilistic diffusion model~\cite{thermo} is a generative model comprised of a forward diffusion process and a reverse diffusion process. The \emph{forward process} is a first-order Markov chain that introduces noise to samples drawn from a data distribution $\bm{p}^1 \sim q(\bm{p}^1)$, while the \emph{reverse process} is a Markov chain that iteratively denoises a noisy input $\bm{p}^T \sim \mathcal{N}(0, I)$. The manner by which the model is trained to add and remove noise allows it to generate samples from the target data distribution $q(\bm{p}^1)$.

In the case of \alg, $\bm{p}^1 = \{ \bm{p}_1^{1}, \bm{p}_2^{1}, \ldots, \bm{p}_k^1\}$ is the set of object poses, where $\bm{p}_i^{1} \in \mathbb{R}^6$ is the SE(3) pose of block $i$, and $q(\bm{p}^1)$ is the distribution over the poses of the blocks that together form a stable stack.\footnote{We normalize poses before applying the diffusion framework. During inference, the generated poses are unnormalized accordingly.} Following DDPM, \alg uses a forward diffusion process that injects noise as%
\begin{equation}
    \tilde{\bm{p}}^t_i = \sqrt{\bar{\alpha}_t} \bm{p}_{i}^{1} + \sqrt{1 - \bar{\alpha}_t} \beps_i,~~ \beps_i \sim \mathcal{N}(\mathbf{0}, \mathbf{I}),
\end{equation}
where $t \sim \operatorname{Unif}\{1, 2, \ldots, T\}$ is the diffusion timesetep that defines the noise scale, $\bar{\alpha}_t$ is a coefficient determined by the noise schedule, and $\beps_i$ is the noise injected in each step. $\bm{p}_{i}^{1}$ is the stable pose of block $i$, $\tilde{\bm{p}}^t_i$ is noise-injected $p_i^1$ at timestep $t$.
\alg then provides the noisy poses $\tilde{\bm{p}}^t_i~(i = 1, 2, \ldots, k)$ to our \emph{denoising} network (the Transformer in Figure~\ref{fig:arch}) $D_\theta$ along with the diffusion timestep $t$, shapes embeddings $\{s_1, \ldots, s_k\}$ and the silhouette of the blocks $S$. This results in the expression
\begin{equation} \label{eq:denoiser}
    \hat{\beps}_{1:k} = D_\theta(\tilde{\bp}_{1:k}, t, s_{1:k}, S),    
\end{equation}
where the notation $X_{1:k}$ is equivalent to $\{X_1, \ldots, X_k\}$,
and $\beps_i$ is a predicted pose noise for the $i$-th block. With the predicted noises, the training objective for a single sample is
\begin{equation}
    \frac{1}{k} \sum_{i=1}^{k} \| \beps_i - \hat{\beps}_i \|^2.
\end{equation}
We sample diffusion timestep $t$ uniformly random from $\{1, 2, \ldots, T\}$ at each training step.

Once the denoising network is trained, the sampling procedure starts with sampling a noisy pose from Gaussian distribution
\[
\tilde{\bp}^{T}_i \sim \mathcal{N}(\mathbf{0}, \mathbf{I}).
\]
From this initial noises, we iterate the following step from $t=T$ to $1$
\begin{align}\label{eq:one-step-denoise}
\tilde{\bp}^{t-1}_{i} = \frac{1}{\sqrt{\alpha}_t}\left( \tilde{\bp}^t_i - \frac{1-\alpha_t}{\sqrt{1 - \bar{\alpha}_t}}\hat{\beps}_i \right) + \sigma_t \bm{z},
\end{align}
where $\beps_i$ is given by Eq. \ref{eq:denoiser} and $\bm{z} \sim \mathcal{N}(\mathbf{0}, \mathbf{I})$ if $t > 1$, otherwise $\bm{z} = \bm{0}$. The resulting $\tilde{\bp}^{1}_{1:k}$ are the generated stable poses.

\subsection{Model Architecture}
Challenging requirements for the model come from the nature of the task that 1) the model must be able to work with a \emph{variable number of} block poses since different stacks use different number and shapes of blocks; and that 2) the model must process inputs from different modalities, including poses, shapes and silhouette that has spatial information.

Our model (Fig.~\ref{fig:arch}) is built upon the Transformer architecture~\citep{NIPS2017_3f5ee243}, which can process input tokens that may originate from different modalities.  To initialize the process, we use a convolutional neural network (CNN) to predict the block list of a structure from that structure's silhouette, shown in Figure~\ref{fig:motivation}.  The CNN converts the single channel mask input into ten channels, followed by thirty residual layers, max pooling, and two fully connected layers. For training we uniquely encode the number of cubes, rectangles, long rectangles and triangles in a structure using an integer index and proceed by training the CNN $C_{\theta}$ with parametrization $\theta$ to model the joint distribution of block counts, represented by the class probability of each index, using the cross-entropy loss
\\
\begin{equation}
    L(D,\theta) = \frac{1}{|D|} \sum_{(S_i, y_i) \in D} -\log\frac{\exp(C_{\theta}(y_i|S_i))} {\sum_{y_k} \exp(C_{\theta}(y_k|S_i))},
\end{equation}
where $D = \{(S_i, y_i)\}$ is our labeled training set, $S_i$ is the structure's silhouette, and $y_i$ is the index corresponding to the structure's block list. This predicted block list serves as one of the inputs to the subsequent steps of our model.

Given a scene that contains a stable stack of $k$ blocks, we extract a list of their poses $\bm{p}_{1:k} \in \mathbb{R}^{k\times 6}$ and shape embeddings $\bm{s}_{1:k} \in \mathbb{R}^{k\times d}$, where $d=512$ (Fig.~\ref{fig:arch}). 
For pose $\bm{p_i}$, we use a six-dimensional pose representation that consists of Cartesian coordinates for translation and exponential coordinates for orientation. The shape embedding $\bm{s}_i$ is retrieved from a codebook that stores unique trainable embeddings for each shape. The poses are projected onto a $d$-dimensional %
space using an MLP applied independently for each object. Similarly, we project the the diffusion timestep $t \in [1, T]$ onto a $d$-dimensional embedding space. The pose, shape, and diffusion timestep embeddings are summed for each object to obtain $k$ object tokens.
In order to handle a variable number of blocks, we use a fixed number of $N$ object tokens to the Transformer encoder and pad the remaining $N-k$ tokens with zero vectors as necessary.%

The silhouette $S$ of the block structure is given as a binary image $I$ of size $64 \times 64$. Following \citet{dosovitskiy2020vit}, we split the binary image $I$ into $c = 16$ patches, each of size $16 \times 16$, and independtly encode each patch using a two-layer MLP to obtain silhouette tokens $i_{1:c} \in \mathbb{R}^{c\times d}$. We add sinusoidal positional embeddings~\citep{NIPS2017_3f5ee243} $\text{Pos}_{1:c} \in \mathbb{R}^{c\times d}$ to the silhouette tokens to retain spatial information. We then sum the patch embeddings and positional encoding together to get $c$ silhouette tokens.

\alg then concatenates the object, padding, and silhouette tokens and feeds them into a six-layer Transformer encoder. The Transformer uses a hidden dimension $d=512$ and expands the representation by the feedforward network with dimension $d_{\text{ff}} = 2048$. In the self-attention layers, we use $h=8$ self-attention heads, where each head has a dimension of $\frac{d}{h} = 64$.

At the last layer of the encoder, \alg linearly projects each contextualized block token back to pose space ($\mathbb{R}^6$) and compute mean squared error (MSE) loss with original noise added to the corresponding pose to conduct supervised learning, following the DDPM framework. Figure~\ref{fig:arch} summarizes this process and architecture. 

In the experiments reported in the paper, \alg uses $N=10$ tokens, $T=50$, and a linear noise schedule from $[10^{-4},\; 0.189 ]$ based on hyperparameter tuning.\

\subsection{Generating Data} \label{subsec:method-data-generation}
\begin{figure}[!t]
    \centering
    \includegraphics[width=0.9\linewidth]{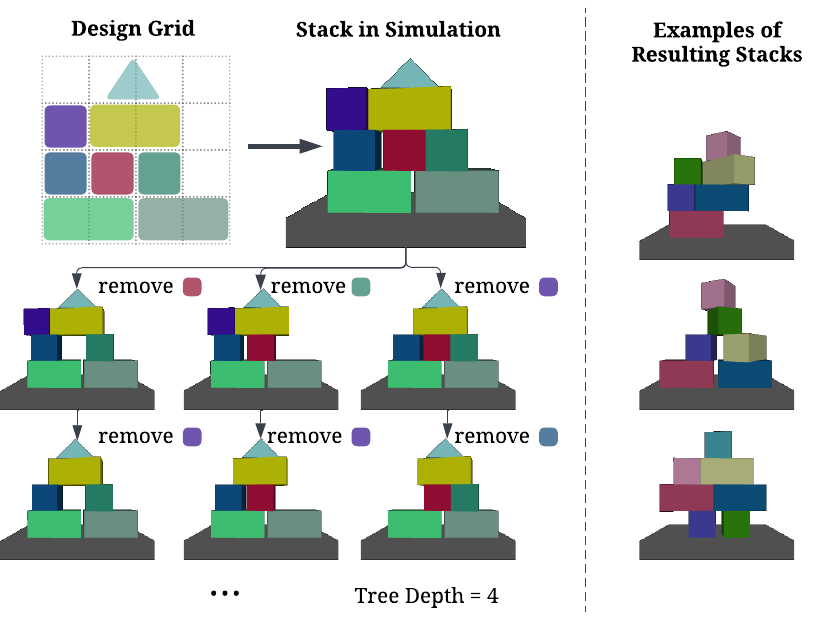} %
    \caption{Our strategy (left) to generate diverse set of stable stacks. After filling the design grid with shapes, we verify stack stability in simulator, and begin removing each block, saving the stable stacks. The right part shows some challenging examples in the dataset.}
    \label{fig:removal}
\end{figure}

To train a model that can generate diverse set of stable block poses, the quality and diversity of the dataset is crucial.
We seek to have an algorithm that synthetically samples various stable block configurations to generate such dataset at scale.
If we place excessive emphasis on diversity of the block stacks, a naive and general approach could be to spawn and drop a randomly selected shape at a random pose in simulation, wait until it settles and repeat this process until a meaningful stack gets constructed in the scene (by checking their height or collisions between blocks, for example).
In the case that this process does not end up in a stack, we could reject it and start over again, repeating the procedure. This could potentially lead to a very general and extremely diverse dataset of block stacking, however, it was found to be inefficient and impractical.

As an alternative, we employ a ``construction by deconstruction'' approach that involves starting with a dense structure comprised of different block shapes, followed by a \emph{block removal} process that involves iteratively removing blocks from the stack until it becomes unstable. While the initial structure is guided by a pre-defined grid, we find that %
the random horizontal displacement and block removal process creates a diverse set of non-trivial structures.

Concretely, we consider a $4 \times 4$ grid that serves as a scaffold for block stack designs. We build the initial dense structure from the bottom up, whereby we attempt to place a randomly chosen block (triangles in the top row only)\footnote{Throughout this paper we consider four different shapes: \{\emph{triangle}, \emph{cube}, \emph{rectangle}, \emph{long rectangle}\}.} in the current row without exceeding a maximum width of four. Once at least three cells in a row are occupied, we move on to the next layer. This results in an initial template of a block stack. We then convert the template to a set of corresponding SE(3) poses for the blocks and add a small amount of noise to their horizontal positions. We then use a simulator to verify that the stack is stable under the influence of gravity, render its front silhouette, and add the set of poses along with the silhouette to the dataset. If the stack falls, we simply reject the design. 
We note that the resulting dataset contains the blocks with slight rotations about the vertical axis, as shown on the right in Figure \ref{fig:removal}. This is due to the inaccuracy of the physics engine, where the blocks keep slightly sliding and rotating randomly while we run forward dynamics and wait for the other part of the stack to be stable. Although not intended, we keep these in the dataset considering that this randomness helps increase the diversity of the block poses.

For each stack of blocks generated as above, we proceed to generate additional data points via \emph{block removal}, whereby we remove blocks whose absence does not collapse the structure. From the initial stack of blocks, as depicted in Figure~\ref{fig:removal}, we try removing each block and simulate the effect on the remaining blocks in the stack. If the stack remains stable, we add the resulting set of block poses and the silhouette to the dataset, and then repeat with another block. We apply this procedure recursively to each stable configuration, removing at most four blocks. We note that the block at the top of the stack is excluded from removal, and thus data samples always have the height of four cubes. Following this procedure, we generate $191$k instances of stable block stacks that we then split into training and test sets using a 9:1 ratio.

\begin{figure}[!tb]
    \centering
    \begin{subfigure}{\columnwidth}
        \centering
        \begin{subfigure}{0.184\columnwidth}
            \includegraphics[width=\textwidth]{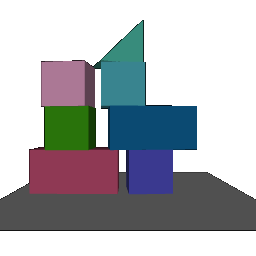}
        \end{subfigure}%
        \hfill%
        \begin{subfigure}{0.184\columnwidth}
            \includegraphics[width=\textwidth]{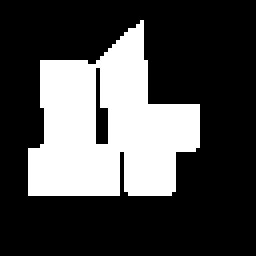}
        \end{subfigure}%
        \rulesep
        \begin{subfigure}{0.184\columnwidth}
            \includegraphics[width=\textwidth]{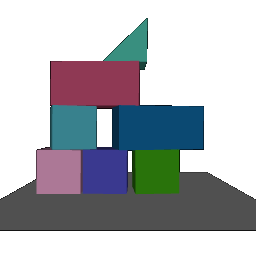}
        \end{subfigure}%
        \hfill%
        \begin{subfigure}{0.184\columnwidth}
            \includegraphics[width=\textwidth]{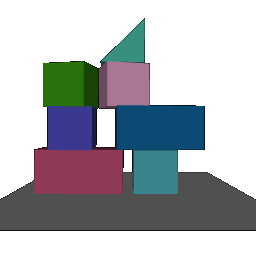}
        \end{subfigure}%
        \hfill%
        \begin{subfigure}{0.184\columnwidth}
            \includegraphics[width=\textwidth]{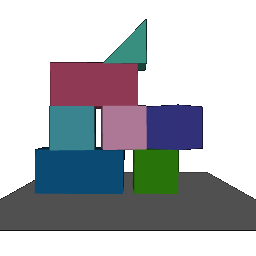}
        \end{subfigure}
    \end{subfigure}
    \vspace{0.5em}
    \begin{subfigure}{\columnwidth}
        \centering
        \begin{subfigure}{0.187\columnwidth}
            \includegraphics[width=\textwidth]{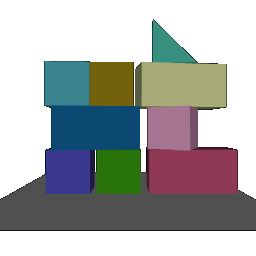}
        \end{subfigure}%
        \hfill%
        \begin{subfigure}{0.184\columnwidth}
            \includegraphics[width=\textwidth]{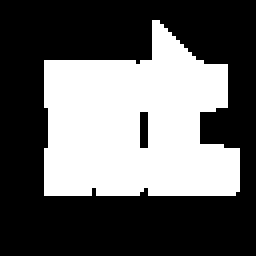}
        \end{subfigure}%
        \rulesep
        \begin{subfigure}{0.187\columnwidth}
            \includegraphics[width=\textwidth]{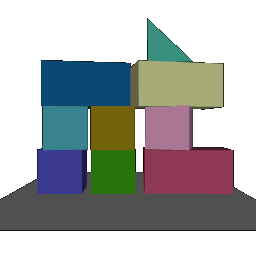}
        \end{subfigure}%
        \hfill%
        \begin{subfigure}{0.187\columnwidth}
            \includegraphics[width=\textwidth]{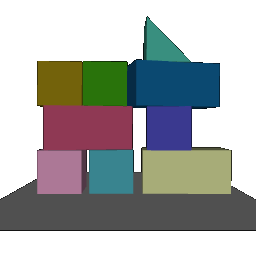}
        \end{subfigure}%
        \hfill%
        \begin{subfigure}{0.187\columnwidth}
            \includegraphics[width=\textwidth]{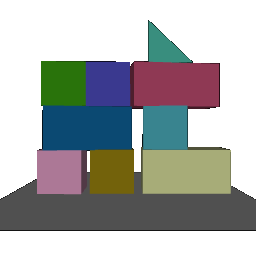}
        \end{subfigure}
    \end{subfigure}
    \caption{A (left) reference (i.e., ground-truth) stack with its silhouette, and (right) a diverse set of structures generated from the silhouette by our model.}
    \label{fig:silhouettes-vs-generated-stack--diversity}
\end{figure}

\section{Experiments}
\label{sec:experiments}

We evaluate the ability of our model to generate a stable configuration of objects that is consistent with a reference input that can take the form of an example of the block structure or a sketch of the desired structure (Fig.~\ref{fig:motivation}). We then present real-world results that involve building different structures using a UR5 robot arm.

\subsection{Evaluation in simulation}
\label{subsec:exp-eval-in-sim}

We evaluate our model using a held-out test dataset. %
Figure~\ref{fig:silhouettes-vs-generated-stack--diversity} visualizes a diverse set of stacks produced by \alg for a single silhouette, demonstrating its capability for multimodal distribution learning.

We aim to evaluate our approach with two metrics: 1) the proportion of block configurations generated by our method that are structurally stable; and 2) the consistency of the generated stacks with the target silhouette. The problem of generating a stable structure from a given block list and silhouette can have multiple solutions, so our evaluation technique samples three sets of block poses for each pair of silhouette and block list in the test set. We compare our method against four baselines: two heuristic baselines, the \emph{Brute-Force Baseline} and the \emph{Greedy-Random Baseline}, and two learning-based baselines, the \emph{Transformer-Regression Baseline} and the \emph{Transformer-VAE Baseline}.

\subsubsection{Brute-Force Baseline}
Given a silhouette and a set of available blocks, this algorithm searches for potential placement poses of each block by maximizing a silhouette alignment score given by silhouette intersection while minimizing a collision penalty between predicted blocks. To achieve a high alignment score, for each block we sample 20 coordinates ${(x_i, 0, z_i)}$ where $x_i$ and $z_i$ are sampled uniformly from $[-3, 3]$ and $\{1,3,5,7\}$, respectively. We perform 20 linear searches from each point along the x-axis in both directions to find poses with optimal alignment and collision measures. This algorithm employs a brute-force search over a discretized space of $k\times 400$ candidate positions, which, although not NP-hard, is computationally intensive.

\subsubsection{Greedy-Random Baseline}
This approach uses a left-to-right, bottom-to-top algorithm operating on the structure silhouette to place blocks. Starting from the lowest layer to the highest, each layer is assigned a fixed height. The algorithm measures the distance of the longest consecutive line of pixels from left to right. It then considers all blocks in the current block list whose width is less than this distance and greedily places the longest one. Since this algorithm is deterministic, we introduce a swap mechanism to add diversity: with a certain probability $\sigma$, the algorithm will swap two adjacent cubes within the same layer with a rectangle elsewhere in the silhouette (since two cubes and a rectangle are of equal length). By controlling the probability $\sigma$, we can adjust the diversity of the generated configurations. This swap mechanism is also applied to the Brute-Force baseline to control its variability.

\begin{figure}[!t]
    \centering
    \includegraphics[width=1.0\linewidth]{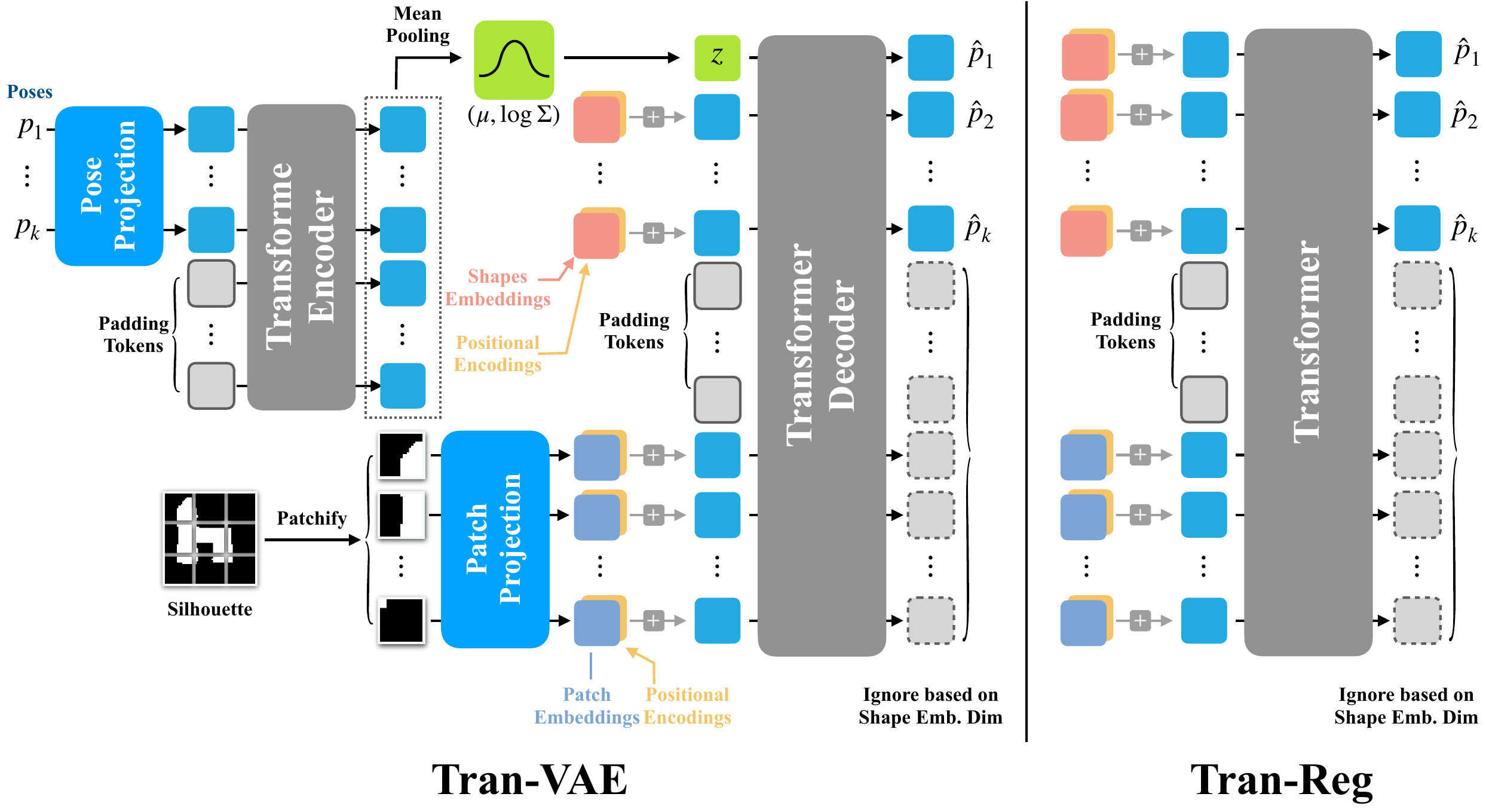} %
    \caption{Our learning-based baselines include (left) Tran-VAE: a Transformer-based VAE model with a two-layer encoder and four-layer decoder, and (right) Tran-Reg: a Transformer-based regression model.}
    \label{fig:baselines}
\end{figure}
\subsubsection{Transformer-Regression Baseline}
We include a Transformer-based regression model (Fig.~\ref{fig:baselines} (right)) as a non-generative, learning-based method. This baseline provides a means to determine the benefits of using a generative model to capture the intrinsically multimodal distribution of block poses conditioned on the silhouette.
To ensure a fair comparison, we train a Transformer with the same number of layers, number of heads, feedforward dimension, and hidden dimension as the Transformer model in \alg. 

\subsubsection{Transformer-VAE Baseline}

The Tran-VAE baseline (Fig.~\ref{fig:baselines} (left)) employs a Transformer-based variational autoencoder with two encoder layers and four decoder layers, matching the total number of layers (six) in \alg's diffusion model. All three learning-based model including \alg using the same batch size and training epoch to make fair comparison.

The transformer-based baselines (Tran-Reg and Tran-VAE) have positional encodings added to the shape embeddings, whereas \alg does not. This is because the diffusion models can differentiate different instances of the same block by their distinct noisy poses.

To quantify the diversity of predicted poses, we analyze the predicted results by obtaining a per-layer block list arranged from left to right. Two constructions are considered distinct if their layered block lists differ. The diversity score is then defined as the number of distinct poses sample poses generated for a given input divided by the total number of samples taken. For example, Figure~\ref{fig:silhouettes-vs-generated-stack--diversity} contains two scenes with average diversity score of $\frac{5}{6}=$ 83.33\%. 

\begin{figure}
    \centering
    \includegraphics[width=1.0\linewidth]{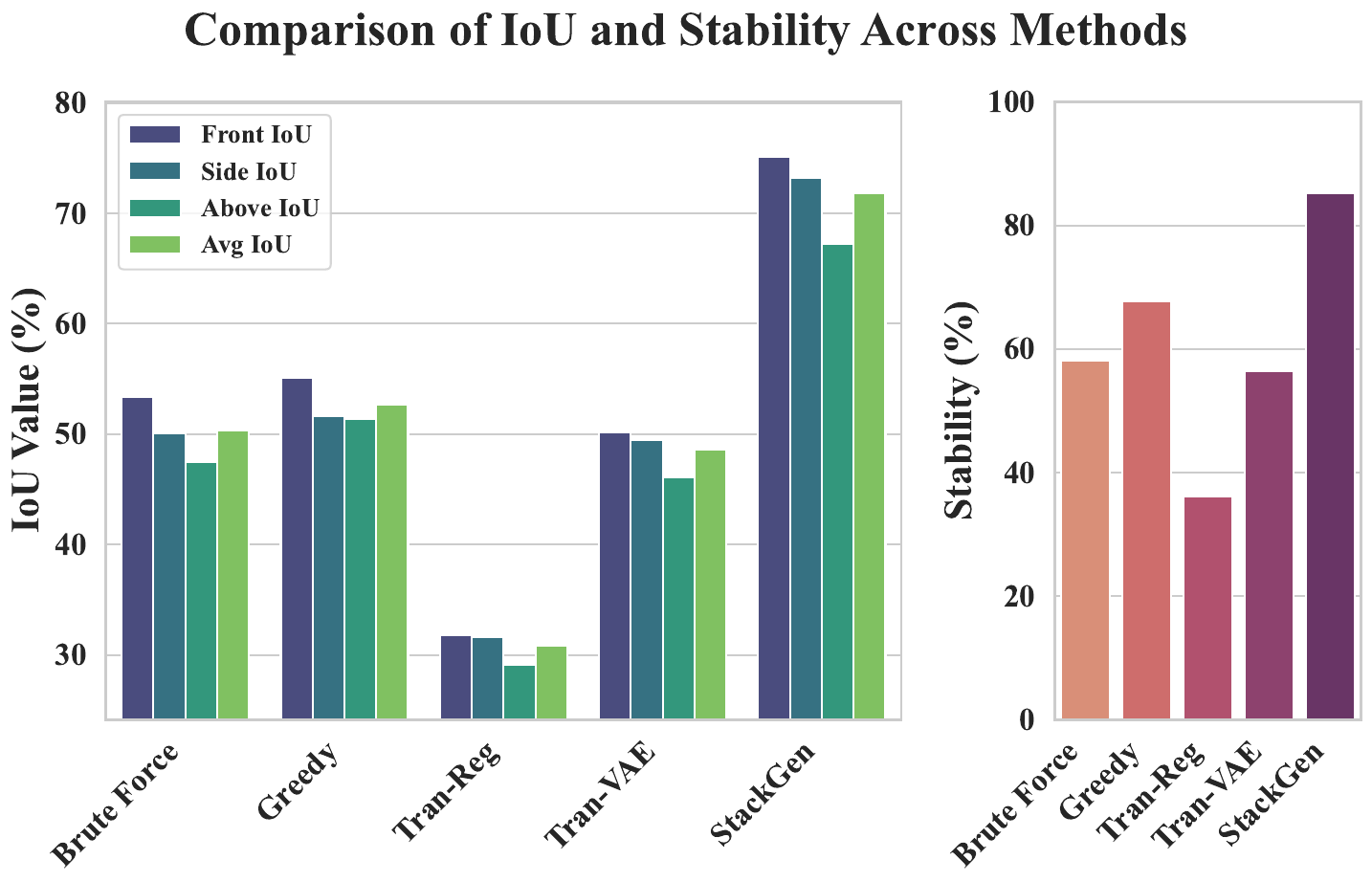}
    \caption{Bar plots that compare \alg with the baselines in terms of (left) IoU and (right) stability. All methods use the CNN-predicted block list.}
    \label{fig:stats_res_fig}
\end{figure}

\begin{figure*}[!t]
    \centering
    \begin{subfigure}{0.48\linewidth}
        \centering
        \includegraphics[height=4.05 cm]{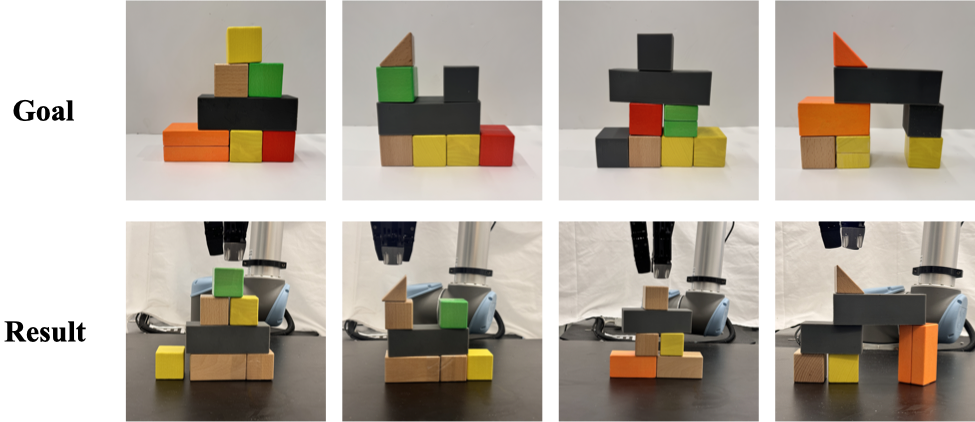}
        \caption{Stack$\rightarrow$stack experiments}
        \label{fig:real-robot-exp-1}
    \end{subfigure}%
    \hfill
    \begin{subfigure}{0.48\linewidth}
        \centering
        \includegraphics[height=4.05 cm]{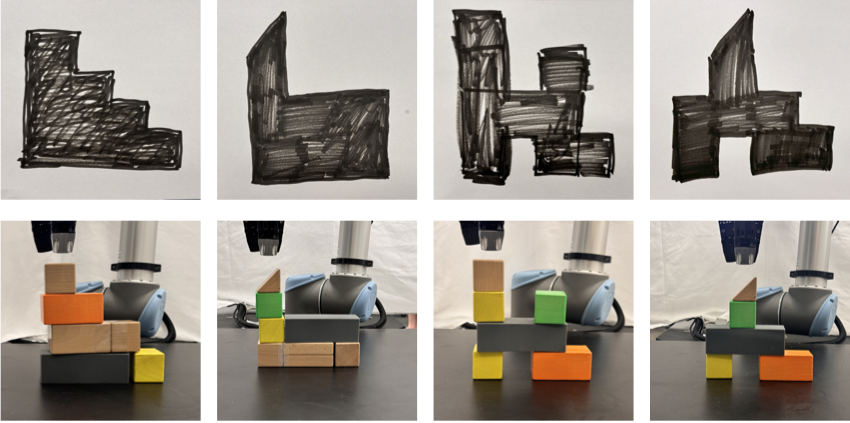}
        \caption{Sketch$\rightarrow$stack experiments}
        \label{fig:real-robot-exp-2}
    \end{subfigure}
    \caption{Examples of various stable 3D structures constructed by a UR5 robot arm based upon goal specifications in the form of \subref{fig:real-robot-exp-1} images and \subref{fig:real-robot-exp-2} sketches of the target structure. Note that \alg seeks to match the silhouette of the input and as a result, the color and type of individual blocks may differ from the reference input.}
    \label{fig:real-stack-vs-generated-and-reconstructed}
\end{figure*}

For stability evaluation, we spawn blocks according to their generated poses, observe their subsequent behavior (i.e., using a simulator for non-real-world experiments), and check whether any of the blocks fall to a layer below where they began, which would lead the sample to be classified as unstable. To assess silhouette consistency, we extract the silhouette of the generated structure after running forward dynamics, compute the intersection over union (IoU) for the silhouettes from three different views (front, side, and top, shown in Figure~\ref{fig:stats_res_fig} left), and then calculate the average IoU across these views. Unstable (collapsed) structures receive an IoU of zero.

\begin{table}[h]
    \label{tab:compare-models} 
    \centering
    {%
    \setlength{\tabcolsep}{7pt}
    \begin{tabular}{lcccr}
        \toprule
        \textbf{Method} & \textbf{Stability} & \textbf{Diversity} & \textbf{Front IoU} & \textbf{Avg IoU}  \\
        \midrule
        \textbf{Brute Force$^\dagger$} & 67.33 & 38.73 & 62.72 & 58.45 \\
        \textbf{Brute Force} & 58.13 & 39.73 & 53.39 & 50.31  \\
        \textbf{Greedy$^\dagger$} & 68.80 & 56.53 & 56.01 & 53.51  \\
        \textbf{Greedy} & 67.67 & 57.53 & 55.13 & 52.72 \\
        \midrule
        \textbf{Tran-Reg$^\dagger$} & 49.73 & 38.80 & 44.57 & 42.97  \\
        \textbf{Tran-Reg} & 36.13 & 46.73 & 31.78 & 30.86  \\
        \textbf{Tran-VAE$^\dagger$} & 71.87 & 56.27 & 65.46 & 63.36   \\
        \textbf{Tran-VAE} & 56.53 & 60.33 & 50.19 & 48.58  \\
        \midrule
        \textbf{{\alg}$^\dagger$} & 84.20 & \textbf{66.87} & 74.39 & 71.31 \\
        \textbf{\alg} & \textbf{85.20} & 64.53 & \textbf{75.08} & \textbf{71.84} \\
        \bottomrule
    \end{tabular}%
    }
    \caption{Comparison of different methods on stability, diversity, and IoU. All results are presented as percentages (\%). The $\dagger$ label indicates models that used the ground-truth block list, those without used the CNN-predicted block list.}
    \label{tab:methods-comparison}
\end{table}

We evaluated all five models using our ground-truth and pretrained CNN block list predictor on 500 scenes, with three samples generated per scene. Since \alg achieved a diversity level of $60.47\%$, we set $\sigma=0.6$ in the Brute-Force and Greedy-Random baselines to match this diversity level. As shown in Table~\ref{tab:methods-comparison} and Figure~\ref{fig:stats_res_fig}, \alg significantly outperforms the baselines in both stability and IoU.

By virtue of being built upon a diffusion model, which is able to represent multimodal distributions, \alg  naturally produces a diverse range of stable designs that meet the silhouette constraints, whereas the heuristic- and learning-based baselines perform noticeably worse in terms of the stability of the resulting stacks, their consistency with the provided sketch (i.e., IoU), and their, perhaps with the exception of the Tran-VAE baseline, their ability to generate diverse designs that match the input silhouette.
Meanwhile, although the brute-force method achieves a relatively high IoU score, it requires evaluating an enormous number of pose combinations, resulting in an inference times that exceed six hours (compared to seconds for \alg). Needless to say, the computational complexity makes them impractical for real-world robotic applications.%

\subsection{Predicted vs.\ Ground-Truth Block Lists}

As a means of evaluating the influence of our CNN-based prediction of the block list, we compare against methods that are provided access to the ground-truth block list.
Using $500$ scenes from the test dataset, we generated three samples per scene and run each model using the CNN-predicted block list $\{\hat{\bm{s}}_{1:k}\}$ and the ground-truth block list $\{\bm{s}_{1:k}\}$. As shown in Table~\ref{tab:methods-comparison} (models annotated with the $\dagger$ were provided the ground-truth block list), the performance of \alg in terms of both stability and IoU differs by no more than $2\%$ between the the use of the CNN-predicted and ground-truth block lists, confirming the utility of the CNN-based predictions for \alg. However, we note the other learning-based methods appear to be highly sensitive to variations in the block list, even though the list is sufficient to generate the stack.

\subsection{Block Stacking in the Real World}
\label{subsec:exp-in-real}

To demonstrate that our method performs well in a real-world environment, we conducted an experiment using toy blocks and a UR5 robotic arm. Our goal was to build a pipeline that operates as follows: first, a user provides a silhouette by either presenting a reference stack of toy blocks or drawing a sketch of their desired structure. After extracting a silhouette from the stack or sketch, our model generates a stable configuration of blocks that matches the provided silhouette. Finally, the UR5 arm assembles the generated stack on a table using real blocks.

\subsubsection{Stack$\rightarrow$stack} In this scenario, a silhouette is extracted from a stack using a simple rig consisting of an RGBD camera (Realsense 435D), toy blocks, and a white background, as shown in Figure~\ref{fig:motivation}. The rig captures a photo of a stack of blocks built by the user then makes a binary silhouette by filtering out background pixels using depth readings and applying a median filter to smooth the silhouette, removing any remaining white pixels, finally resizing and pasting the result onto a $64\times 64$ canvas.

\subsubsection{Sketch$\rightarrow$stack} 
In this case,
we use the
camera to capture a hand-drawn sketch from a user (Figure~\ref{fig:motivation}). 
It is converted into a binary image, smoothed using a median filter, and a bounding box with a $4\times 4$ grid is put around it. We then compute the occupancy 
to identify whether each grid cell is fully or partially occupied (e.g., a triangle).

With the extracted silhouettes, we use our pretrained CNN to predict the block list.\footnote{For the sketch$\rightarrow$stack example, we employ a heuristic method rather than the CNN to identify the block list.} The diffusion model then generates candidate block poses. For each block in a set of generated poses, the UR5 arm executes a pick-and-place operation to position the block at its corresponding pose. The execution sequence is set greedily from left to right and bottom up.

Out of the eight cases we tested, this pipeline successfully built all of the stacks stably, with only minor discrepancies relative to the original silhouettes (considering the error of block initial position). However, we note that this does not imply that our system is flawless. As discussed in Section~\ref{subsec:exp-eval-in-sim}, the model can sometimes generate unstable block configurations. Nonetheless, in these real-world experiments, the success rate indicates that the model is robust enough to handle potentially out-of-distribution silhouettes effectively.
\section{Conclusion}

In this paper, we presented a new approach that enables robots to reason over the 6-DoF pose of objects to realize 
a stable 3D structure. Given a dataset of stable structures, \alg learns a distribution over the SE(3) pose of different object primitives, conditioned on a user-provided silhouette of the desired structure. At inference time, \alg generates a diverse set of candidate compositions that align with the silhouette while ensuring physical feasibility. 

We conducted experiments in a simulated environment and showed that
our approach effectively generates stable structures following a user-provided silhouette, without modeling physics explicitly.
Further, we deployed our approach in a real-world setting, demonstrating that the method effectively and reliably generates stable and valid block structures in a data-driven manner, bridging the gap between visual design inputs and physical construction.

{\small
\bibliographystyle{IEEEtranN}
\bibliography{references}
}

\end{document}